\documentclass[10pt]{article}

\usepackage{graphicx}
\usepackage{url}
\usepackage{amsmath, amssymb, amsfonts}
\usepackage{xcolor}
\usepackage{threeparttable}
\usepackage{tabularx}
\usepackage{booktabs}
\usepackage{cite}
\usepackage{subfig}

\DeclareMathOperator*{\argmin}{arg\,min}

\title{\LARGE\bfseries
A Tightly Coupled Magnetic-Inertial SLAM System for Field Robots
}

\author{Chuan Huang \and Tongxing Peng}

\begin{document}

\maketitle

\begin{abstract}
Field robots operating in mines, tunnels, and disaster areas require reliable localization despite GNSS denial and degraded camera and LiDAR measurements 
caused by dust, smoke, or poor illumination. Missions involving both ground and aerial robots further motivate complementary sensing modalities that do not depend on visibility, preinstalled
infrastructure, or wheel contact. This letter presents MISLAM, a tightly coupled magnetic-inertial simultaneous localization and mapping system using an IMU and a nine-magnetometer array, without relying on external odometry or platform-specific motion assumptions. 
The system combines the inertial navigation model with local polynomial magnetic-field models for short-term motion estimation and a global Gaussian process model for drift correction using an error-state Kalman filter. 
Offline evaluation on three indoor datasets collected with a handheld rig shows reductions in total endpoint closure error of 45.7\%–78.0\%
relative to magnetic-inertial odometry. Under poor illumination, MISLAM maintains tracking 
after visual-inertial odometry fails, achieving a horizontal closure error of 0.332 m, although vertical drift remains a limitation. These results support magnetic-inertial SLAM as a complementary localization modality for robotic sensor suites operating under degraded visual conditions.
\end{abstract}

\section{Introduction}

Reliable localization is a prerequisite for field robots exploring mines,
tunnels, and disaster areas. These environments often deny access to global
navigation satellite systems (GNSS), while poor illumination, dust, and smoke
can degrade camera- and LiDAR-based localization~\cite{Ebadi2024Present}.
These challenges motivate complementary sensing modalities that provide
motion information without relying on visibility, preinstalled
infrastructure, or wheel contact. Such capabilities are relevant to both
ground and aerial robots and can support localization when other sensors
become unreliable.

Indoor magnetic fields have been recognized as a promising environmental feature for localization 
because of their spatial inhomogeneity and temporal stability. Ferromagnetic structures and objects create spatial variations that can remain sufficiently stable to serve as location-dependent signatures~\cite{Li2012Magnetic}.
Furthermore, magnetic-field-based localization does not require capturing identifiable visual information, avoiding the associated image-privacy concerns. These properties have motivated sustained research over the past two decades, 
spanning magnetic-field fingerprinting and model-based localization methods~\cite{Vissiere2007magnetometers,Vallivaara2010Simultaneous,Kok2018MagSlam,zmitri2019improving,Viset2022EKF,pavlasek2023magnetic,
huang2023mains,ZHANG2023MagODO,Kok2024Online,Shen2024IDFMFL,Kuang2025CrowdMagMap,huang2026inertialmagneticslamsystems,Vallivaara2026Saying,Tongxing2026MAVIS}.

Magnetic fingerprinting typically matches measurements to locations within a pre-surveyed map. However, surveying an unknown environment in advance can be time-consuming and is often infeasible prior to a field robotics mission.
Magnetic-field SLAM addresses this issue by estimating the trajectory and field map jointly. Representative methods~\cite{Kok2018MagSlam,Viset2022EKF,Vallivaara2026Saying} use
Gaussian processes (GPs) to model the magnetic field along the trajectory and constrain the drift accumulated during pose propagation. Despite these advances, existing magnetic SLAM methods still rely on accurate external odometry---typically from visual-inertial odometry (VIO) or pedestrian dead reckoning---which limits their applicability when primary odometry sensors degrade or the platform's motion pattern is incompatible with the assumptions of the odometry method.
Achieving self-contained magnetic-field SLAM without external odometry or restrictive motion assumptions presents two main challenges: 1) High computational complexity: Extracting odometry directly from magnetic field measurements requires high-resolution field models capable of capturing sub-meter spatial variations. As reported in~\cite{skog2025OnTheConnection}, 50 to 500 basis functions per square meter are needed to support accurate localization.
2) Degenerate spatial sampling: Spatially proximate measurements often yield overconfident predictions of the magnetic field at nearby locations, particularly when entering unexplored areas~\cite{Vallivaara2010Simultaneous,lee2020magslam}. Mitigating this issue requires enforcing a minimum guard distance or time interval between map updates, which further complicates high-resolution modeling.


Magnetometer arrays enable an alternative, local modeling of the magnetic fields across the array. Because the array covers a relatively small volume, accurate yet computationally efficient local field models can be constructed.
Moreover, because the relative positions of the magnetometers are fixed and known, the local field model can be fitted and evaluated from a single multi-sensor snapshot with lower uncertainty than fitting single-magnetometer measurements along an estimated trajectory. 
This type of modeling enables magnetic odometry inspired by optical flow~\cite{skog2021magnetic}. Magnetic-inertial
odometry (MIO) systems, including MAINS~\cite{huang2023mains} and other array-based methods~\cite{Vissiere2007magnetometers,zmitri2020magnetic,Li2025SCEKF,liu2026fss}, fuse array measurements with inertial measurements to bound velocity and position drift. Although these methods are self-contained and require neither external odometry nor motion constraints, existing MIO systems do not maintain a persistent global field map. Consequently, they cannot exploit loop closures when revisiting mapped areas, leaving long-term navigation drift uncorrected.

Building on these developments, this letter presents MISLAM, a tightly coupled magnetic-inertial SLAM system that combines the complementary strengths of local polynomial field modeling and global GP mapping using a single error-state Kalman filter (ESKF). The main contributions of this letter are as follows:

\begin{itemize}
    \item We develop a self-contained magnetic-inertial SLAM system using an IMU and a nine-magnetometer array that operates without external odometry or restrictive motion constraints.

    \item We develop an ESKF update strategy that validates local field references using chi-square tests, retires each reference after one update attempt, and selects global GP keyframes based on the number of accepted local updates.

    \item We evaluate MISLAM against a state-of-the-art MIO system (MAINS~\cite{huang2023mains}) and a VIO system (OpenVINS~\cite{Geneva2020ICRA}) across three indoor datasets collected with a handheld sensor 
    rig under varying revisit topologies and lighting conditions. 

\end{itemize}

\section{Overview}
\begin{figure*}[tb!]
    \centering
    \includegraphics[width=\textwidth]{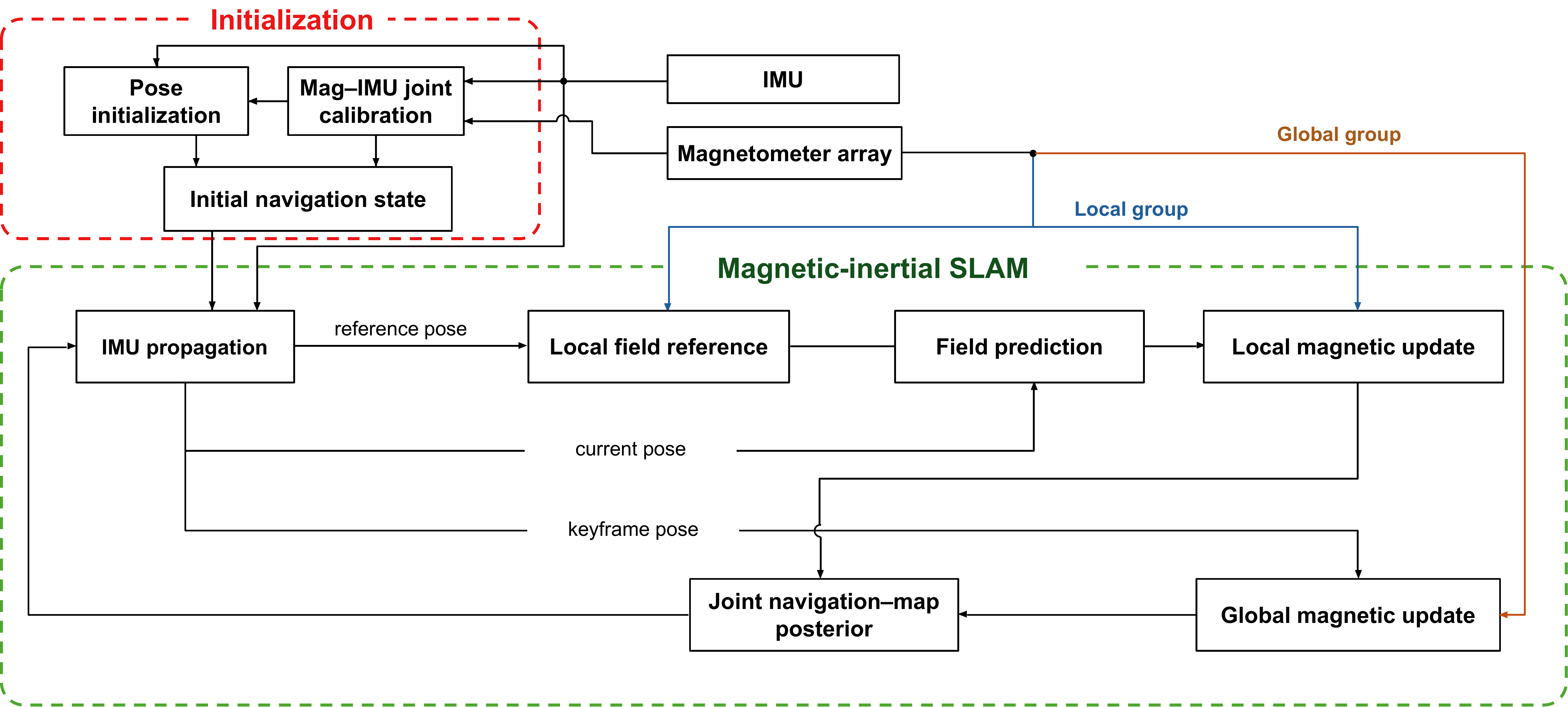}
    \caption{System overview of the tightly coupled magnetic-inertial SLAM architecture.}
    \label{fig:system-overview}
\end{figure*}

The proposed MISLAM system fuses inertial and magnetic measurements to jointly estimate the trajectory and the magnetic-field map. 
The system uses a sensor rig that consists of an array of $N$ magnetometers and an IMU. The processing stages---sensor calibration, stationary initialization, inertial propagation,
and local and global magnetic updates---are shown in Fig.~\ref{fig:system-overview} and detailed in Sections~\ref{sec:initialization} and~\ref{sec:magnetic inertial slam}. Here, we present the main concepts and notation.

After proper sensor calibration and filter initialization, IMU measurements propagate the position,
velocity, and attitude. Magnetic measurements from two disjoint groups—the local group \(L\) and the global group \(G\)—reduce the propagation error through the local and global field representations, respectively.
A local model captures the spatial field variation
around a recent reference pose and provides short-term relative
motion information. A global model retains
magnetic information across the explored environment, allowing
measurements in previously mapped regions to reduce accumulated
navigation drift. 

Throughout this letter, $W$ denotes the navigation world frame, $W_{\mathrm{cal}}$ denotes the calibration world frame, $b$ denotes
the body frame coinciding with the IMU frame, and $\mathcal M_i$
denotes the coordinate frame of the $i$th magnetometer.
The world-frame origin is placed at the initial IMU position.
Its $z$-axis points upward, opposite to gravity, such that
$g^W=[0,0,-g]^\top$. The frame \(W_{\mathrm{cal}}\) has its \(z\)-axis pointing upward and its \(y\)-axis aligned with the horizontal component of the calibration magnetic field. A rotation matrix $R_A^B$
maps vectors from frame $A$ to frame $B$; in particular,
$R_k=R_{b_k}^W$ denotes the body-to-world rotation at time $k$.
The fixed displacement from the IMU origin to magnetometer $i$,
expressed in the body frame, is denoted by $d_i$.

\section{System Initialization}\label{sec:initialization}

\subsection{Magnetometer-IMU Joint Calibration}
Accurate magnetic-inertial fusion requires consistent magnetic
measurements expressed relative to the IMU frame. Magnetometer
measurements are affected by hard-iron offsets, soft-iron distortions,
scale-factor errors, and nonorthogonal sensing axes. Mounting errors
also introduce misalignment between each magnetometer and the IMU.
We account for these effects using the joint calibration method
in~\cite{Huang2026Joint}.


Calibration data are collected by slowly rotating the assembled sensor rig through a wide range of orientations in an open outdoor area, where the magnetic field is approximately constant. Translational acceleration is kept
small so that the accelerometer measurements are dominated by gravity. Within this subsection, let \(R_k \triangleq R_{b_k}^{W_{\mathrm{cal}}}\) denote the rotation from the IMU frame to the calibration world frame. The orientation dynamics and measurements of the sensor rig are modeled as
\begin{subequations}
\label{eq:calibration-model}
\begin{align}
 R_{k+1}
 &=R_k\operatorname{Exp}\!\left(
 [(\tilde{\omega}_k-o^\omega-e_k^\omega)\Delta t]_\times
 \right),\\
 \tilde{a}_k
 &=-R_k^\top g^{W_{\mathrm{cal}}}+o^a+e_k^a,\\
 \tilde{m}_k^i
 &=C_iR_k^\top B_{\mathrm{cal}}^{W_{\mathrm{cal}}}(\iota)+o^{m_i}+e_k^{m_i},
 \; i=1,\ldots,N,
\end{align}
\end{subequations}
where $[\cdot]_\times$ denotes the operator that maps $\mathbb{R}^3$ to the Lie algebra $\mathfrak{so}(3)$. The variables $\tilde{\omega}_k$ and $\tilde{a}_k$ denote the gyroscope and accelerometer measurements, respectively, with corresponding biases $o^\omega$ and $o^a$. The sampling interval is denoted by $\Delta t$. For the $i$th magnetometer, $\tilde{m}_k^i$ and $o^{m_i}$ denote its measurement and bias, respectively. 
The vector $B_{\mathrm{cal}}^{W_{\mathrm{cal}}}(\iota)=\big[0 \; \cos(\iota) \; -\sin(\iota)\big]^{\top}$ denotes the \textit{normalized} magnetic field expressed in the calibration world frame, where $\iota$ is the dip angle. The fixed matrix $ C_i=D^iR_b^{\mathcal M_i}$
combines the intrinsic distortion matrix $D^i$ and the rotation $R_b^{\mathcal M_i}$ from the IMU frame to the $i$th magnetometer frame. For all sensors, the terms $e_k^{(\cdot)}$ denote measurement noise, which is assumed to be white, zero-mean Gaussian with known covariance $\Sigma_{(\cdot)}$.


Let $\theta \triangleq (o^a, o^\omega, \{C_i, o^{m_i}\}_i, \iota)$ denote the calibration parameters and $R_{0:T}$ the orientation trajectory. The calibration parameters and orientation trajectory are jointly estimated by minimizing the following maximum-likelihood objective:
\begin{subequations} \label{eq: calibration}
    \begin{equation} 
        (\hat{\theta}, \hat{R}_{0:T})^{\text{\tiny{ML}}} \triangleq \argmin_{\theta, R_{0:T}} L(\theta,R_{0:T})      
    \end{equation}
where
    \begin{equation}
        \begin{split} 
            L(\theta,&R_{0:T})= \sum_k\|\tilde{a}_k+R_k^{\top} g^{W_{\mathrm{cal}}} - o^a\|^2_{\Sigma_a}\\
            &+  \sum_k\|  \tilde{\omega}_k - \frac{\operatorname{Log}(R_k^{\top} R_{k+1})}{\Delta t}  - o^\omega \|^2_{\Sigma_\omega}\\
            & + \sum_i \sum_k\|\tilde{m}_k^i- C_i R_k^{\top} B_{\mathrm{cal}}^{W_{\mathrm{cal}}}(\iota) - o^{m_i}\|^2_{\Sigma_{m^i}}.
        \end{split}
    \end{equation}
\end{subequations}
Here $\operatorname{Log}:SO(3)\to\mathbb R^3$ denotes the rotation-vector logarithm and $\|r\|_\Sigma^2\triangleq r^\top\Sigma^{-1}r$.
The estimated calibration parameters $\hat{\theta}$ are then used in the subsequent pose initialization and state estimation stages.


\subsection{Stationary Pose Initialization}
After joint magnetometer-IMU calibration, the rig is held stationary for initialization.
The position and velocity are set to zero. The initial attitude is derived from accelerometer measurements. Let $\bar a_0$ denote the mean
bias-corrected accelerometer measurements during this phase. Under stationary conditions,
$\bar a_0\approx-R_0^\top g^W$. We parameterize the initial attitude as
$R_0=R_z(\psi_0)R_y(\vartheta_0)R_x(\phi_0)$.
The initial roll and pitch are therefore~\cite{pedley2013tilt}
\begin{equation}
 \begin{aligned}
 \phi_0
 &=\operatorname{atan2}(\bar a_{0,y},\bar a_{0,z}),\\
 \vartheta_0
 &=\operatorname{atan2}\!\left(
 -\bar a_{0,x},
 \sqrt{\bar a_{0,y}^2+\bar a_{0,z}^2}
 \right).
 \end{aligned}
 \label{eq:initial-tilt}
\end{equation}
By convention, we set $\psi_0=0$, so that the initial heading of the sensor rig defines the horizontal orientation of the world frame.

\section{Magnetic-Inertial SLAM System}\label{sec:magnetic inertial slam}
\label{sec:mi-slam}

The proposed MISLAM system combines inertial propagation and observations of indoor magnetic fields, on both local and global scales, in a single state-space model and uses the
error-state Kalman filter~\cite{Joan2017Quaternion} for state estimation. After initialization, each incoming IMU sample propagates the navigation state and joint
covariance of the navigation state, global magnetic field model coefficients, and the reference pose when available. When an eligible local magnetic reference is available,
a local magnetic update is attempted. The estimator then evaluates the keyframe selection criteria. At selected
keyframes, a global magnetic update is attempted. The online recursion is summarized as follows:
\begin{enumerate}
 \item Propagate the navigation state and joint covariance with the IMU.
 \item Create or validate a local field reference, or attempt an update using an eligible reference. Retire the reference after one update attempt, whether the update is accepted or rejected.
 \item Check the accepted local magnetic update count to determine whether a new keyframe is selected.
 \item At the selected keyframe, attempt the global magnetic update.
 \item Output the current navigation estimate and prepare the joint state
       for the next sample.
\end{enumerate}
The following subsections describe the state estimation algorithm. 
Unless otherwise indicated, state variables denote nominal estimates. Superscripts minus and plus denote the prior and posterior for the current update.

\subsection{Joint State and Error State}
\label{sec:joint-state}

The persistent state $x_k^{\mathrm{p}}$ consists of the navigation state $ x_{n,k}$ and the global
magnetic-field model coefficients $u_k$:
\begin{equation}
 \begin{aligned}
x_k^{\mathrm{p}}&\triangleq(x_{n,k},u_k),\\
 x_{n,k}&\triangleq(p_k,v_k,R_k,o_k^a,o_k^\omega).
 \end{aligned}
 \label{eq:permanent-state}
\end{equation}
Here $p_k$ and $v_k$ are the IMU position and velocity in the world
frame. The vector $u_k$ contains the magnetic field
model coefficients defined in \eqref{eq:gp-field}. 

The corresponding error state is
\begin{equation}
 \begin{aligned}
 \delta x_k^{\mathrm{p}}
 &\triangleq[\delta x_{n,k}^\top,\delta u_k^\top]^\top,\\
    \delta x_{n,k}
 &\triangleq[\delta p_k^\top,\delta v_k^\top,\delta\varphi_k^\top,
       \delta o_k^{a\top},\delta o_k^{\omega\top}]^\top.\\
 \end{aligned}
 \label{eq:permanent-error}
\end{equation}
Vector errors are additive, while the three-dimensional attitude error
is defined by the world-frame left perturbation
$R_k^{\mathrm{true}}=\operatorname{Exp}([\delta\varphi_k]_\times)R_k$.

To estimate relative motion from measurements at different epochs, the filter augments the
persistent state with a temporary local-reference pose
$\xi_r=(p_r,R_r)$, which is a clone of a past navigation
pose and contributes six error coordinates,
$\delta\xi_r=[\delta p_r^\top,\delta\varphi_r^\top]^\top$,
with the same attitude-error convention. When the temporary pose exists,
the augmented state and its error state are
\begin{equation}
 \begin{aligned}
 x_k&=(x_k^{\mathrm{p}},\xi_r),\\
 \delta x_k&=[(\delta x_k^{\mathrm{p}})^\top,
              \delta\xi_r^\top]^\top.
 \end{aligned}
 \label{eq:joint-error}
\end{equation}
The local-reference
block is omitted when no local reference is active. The joint covariance
$P_k=\text{cov}(\delta x_k)$ includes all cross-covariances
between the navigation errors, coefficient errors, and local-reference
pose errors. Pose augmentation and removal are described in
Section~\ref{sec:joint-correction}.

The local magnetic field model coefficients are fitted separately
and are not included in the joint state. 

\subsection{Inertial Propagation}
\label{sec:imu-propagation}
Define the bias-corrected IMU measurements
$f_k=\tilde a_k-o_k^a$ and
$\omega_k=\tilde\omega_k-o_k^\omega$.
Over a sampling interval $\Delta t$, the nominal navigation state
is propagated as
\begin{equation}
 \begin{aligned}
 p_{k+1}^-&=p_k+v_k\Delta t
       +\tfrac12(R_kf_k+g^W)\Delta t^2,\\
 v_{k+1}^-&=v_k+(R_kf_k+g^W)\Delta t,\\
 R_{k+1}^-&=R_k
       \operatorname{Exp}([\omega_k\Delta t]_\times).
 \end{aligned}
 \label{eq:ins-propagation}
\end{equation}
The IMU biases, $o_k^a$ and $o_k^\omega$, are modeled by two independent random walk processes. Linearizing the inertial error dynamics and applying Van Loan
discretization yields the navigation transition $F_{n,k}$ and process
covariance $Q_{n,k}$; for details see~\cite[Ch. 7]{Joan2017Quaternion}. Map coefficients and pose clones have identity
dynamics and no process noise. Thus, with identity and zero blocks sized
to the active static states,
\begin{equation}
 \begin{aligned}
 F_k&=\operatorname{blkdiag}(F_{n,k},I),\\
 P_{k+1}^-&=F_kP_k^+F_k^\top+
                   \operatorname{blkdiag}(Q_{n,k},0).
 \end{aligned}
 \label{eq:joint-propagation}
\end{equation}

\subsection{Magnetic Models and Updates}

\subsubsection{Local Magnetic Model and Update}
\label{sec:local-magnetic-updates}
Within a small region around a reference pose, the local magnetic field is
represented by the polynomial model~\cite{skog2018magnetic}, which is derived from a polynomial scalar potential and
constrained to satisfy the divergence-free
condition. For a polynomial model of degree $n$, the resulting
linear representation has $N_p=n^2+4n+3$ coefficients and is of the following form:
\begin{equation}
 B(s)=\Phi(s)\eta_r,\qquad
 \Phi(s)\in\mathbb R^{3\times N_p},\quad \eta_r\in\mathbb R^{N_p},
 \label{eq: polynomial model}
\end{equation}
where $s$ and $B(s)$ denote the position and magnetic field vector, both expressed in the body frame at reference
epoch $r$, and $\eta_r$ denotes the local
model coefficients.

Stacking the measurements of the local group $L$ into $y_r$ gives the linear
model $y_r=A\eta_r+o^m+e_r$, with blocks
$A_i=C_i\Phi(d_i)$, $o^m_i=o^{m_i}$ for $i\in L$. The vector $e_r$ includes the sensor and modeling errors.
A weighted least-squares fit yields
\begin{equation}
      \hat\eta_r=P_{\eta_r}A^\top\Sigma_{\mathrm{fit}}^{-1}(y_r-o^m),
 \label{eq:local-fit}
\end{equation}
where $P_{\eta_r} \triangleq  \text{cov}(\hat\eta_r)
 =(A^\top\Sigma_{\mathrm{fit}}^{-1}A)^{-1}$ and $\Sigma_{\mathrm{fit}}\triangleq \text{cov}(e_r)$.
 
A reference is accepted only if its normalized fit
residual passes a chi-square test,
\begin{equation}
r_{\mathrm{fit}}^\top\Sigma_{\mathrm{fit}}^{-1}r_{\mathrm{fit}}
 \leq \chi^2_{3N_L-N_p,\,1-\alpha_{\mathrm{fit}}}.   
\end{equation}
Here $ r_{\mathrm{fit}}\triangleq y_r-A\hat\eta_r-o^m$ denotes the fit residuals and $N_L\triangleq|L|$ denotes the number of magnetometers in $L$. The accepted reference's pose is then cloned into the
joint state.

At a later epoch $k$, define the relative rotation and translation
\begin{equation}
R_{rk}
 \triangleq R_{b_k}^{b_r},\qquad
 t_{rk}=R_r^\top(p_k-p_r).
 \label{eq:local-relative-pose}
\end{equation}
The position of the $i$th magnetometer in the reference frame and its
measurement model are 
\begin{equation}
 \begin{aligned}
 s_{i,rk}&=t_{rk}+R_{rk}d_i,\\
 y_{i,k}&=h_{i,k}^{\mathrm{loc}}(x_{k};\eta_r)+e_{i,k}^{\mathrm{loc}},\\
 h_{i,k}^{\mathrm{loc}}(x_{k}; \eta_r)
 &\triangleq C_iR_{rk}^\top\Phi(s_{i,rk})\eta_r+o^{m_i}.
 \end{aligned}
 \label{eq:local-observation}
\end{equation}
Here $e_{i,k}^{\mathrm{loc}}$ includes sensor noise and local model \textit{extrapolation} errors. For the stacked measurements and an estimated $\hat \eta_r$, the effective noise covariance is
\begin{subequations}
    \begin{equation}
    \Sigma_{\mathrm{loc},k}=\Sigma_{L,k}^{\mathrm{loc}}+J_{\eta,k}P_{\eta_r}J_{\eta,k}^{\top},
    \end{equation}
    where
    \begin{equation}        
    \begin{aligned}
        \Sigma_{L,k}^{\mathrm{loc}}
  &=\operatorname{cov}(e_{L,k}^{\mathrm{loc}}),\\
e_{L,k}^{\mathrm{loc}}
  &\triangleq[(e_{i_1,k}^{\mathrm{loc}})^\top,\ldots,
    (e_{i_{N_L},k}^{\mathrm{loc}})^\top]^\top,
    \end{aligned}
    \end{equation}
    and 
    \begin{equation}
        J_{\eta,k} = \begin{bmatrix}
            C_{i_1}R_{rk}^\top\Phi(s_{i_1,rk})\\
            \vdots\\
            C_{i_{N_L}}R_{rk}^\top\Phi(s_{i_{N_L},rk})\\
        \end{bmatrix}.
    \end{equation}
\end{subequations}

A reference is eligible for a local magnetic update after a minimum age, \textit{minRefAge}. It is
retired when its maximum age, \textit{maxRefAge}, is exceeded. Eligible
measurements undergo the innovation test in
Section~\ref{sec:joint-correction}. After one update attempt,
accepted or rejected, the reference is retired; a new reference can be
formed at a later epoch.

\subsubsection{Global Magnetic Model and Update}
\label{sec:global-magnetic updates}

The global magnetic field model represents the world-frame magnetic field. Following the reduced-rank scalar-potential
Gaussian process representation used in magnetic
SLAM~\cite{solin2018modeling}, we write
\begin{equation}
 B^W(s)=b_0+\sum_{j=1}^{M}a_j\nabla\phi_j(s)=W(s)u,
 \label{eq:gp-field}
\end{equation}
where $b_0$ denotes the background magnetic field,
$u=[b_0^\top,a_1,\ldots,a_M]^\top$, and
$W(s)=[I_3,\nabla\phi_1(s),\ldots,\nabla\phi_M(s)]$.
The functions $\phi_j$ are normalized Laplacian eigenfunctions
on a fixed rectangular domain $\Omega$, ordered by increasing
eigenvalue $\lambda_j$. A squared-exponential potential kernel
gives the prior
\begin{equation}
 \begin{aligned}
 u&\sim\mathcal N(0,P_u^0),\quad
 P_u^0=\operatorname{blkdiag}(\sigma_b^2I_3,\Lambda),\\
 \Lambda_{jj}
 &=\sigma_p^2(2\pi\ell^2)^{3/2}
       \exp(-\ell^2\lambda_j/2),
 \end{aligned}
 \label{eq:gp-prior}
\end{equation}
where $\sigma_b$ is the prior standard deviation of each background
field component, and $\ell$ and $\sigma_p$ are the potential
length scale and standard deviation. The domain is specified
before processing and remains fixed during navigation. Its
extent and the number of basis functions determine the spatial
detail represented by the map.

For magnetometer $i\in G$, the sensor position and
its measurement model are
\begin{equation}
 \begin{aligned}
 s_{i,k}^W&=p_k+R_kd_i,\\
 y_{i,k}&=h_{i,k}^{\mathrm{GP}}(x_{k})+e_{i,k}^{\mathrm{GP}},\\
 h_{i,k}^{\mathrm{GP}}(x_{k})
 &\triangleq   C_iR_k^\top W(s_{i,k}^W)u_k+o^{m_i},
 \end{aligned}
 \label{eq:gp-observation}
\end{equation}
where $e_{i,k}^{\mathrm{GP}}$ includes sensor noise and an allowance for GP approximation error. For the stacked measurements, the effective noise covariance is
    \begin{equation}        
    \begin{aligned}
        \Sigma_{k}^{\mathrm{GP}}
  &=\operatorname{cov}(e_{k}^{\mathrm{GP}}),\\
e_{k}^{\mathrm{GP}}
  &\triangleq[(e_{i_1,k}^{\mathrm{GP}})^\top,\ldots,
    (e_{i_{N_G},k}^{\mathrm{GP}})^\top]^\top.
    \end{aligned}
    \end{equation}
Here $N_G\triangleq|G|$ denotes the number of magnetometers in $G$.

The global magnetic update is attempted at selected keyframes.
The initial epoch is a keyframe. Subsequently, a new keyframe is selected
after a prescribed number $N_{\text{\tiny{acc}}}$
of local updates have been accepted since the previous
keyframe.

\subsection{Filter Update and Cloned-Pose Management}
\label{sec:joint-correction}
For innovation
$\nu=y-h(\hat x)$, measurement model Jacobian $H$, and effective noise covariance $\Sigma$,
\begin{equation}
 \begin{aligned}
 S&=HP^-H^\top+\Sigma,\qquad K=P^-H^\top S^{-1},\\
 \delta\hat x&=K\nu,\\
 \widetilde P^+&=(I-KH)P^-(I-KH)^\top+K\Sigma K^\top.
 \end{aligned}
 \label{eq:joint-correction}
\end{equation}
The update is accepted only when
$\nu^\top S^{-1}\nu\leq\chi^2_{d,1-\alpha}$, where $d$ is the
innovation dimension and $\alpha$ is the chosen rejection probability.
Vector corrections are additive; the current and cloned attitudes are
corrected by left multiplication with
$\operatorname{Exp}([\delta\hat\varphi]_\times)$. The covariance is
then reset as $P^+=J_{\mathrm{reset}}\widetilde P^+J_{\mathrm{reset}}^\top$,
using the $SO(3)$ left Jacobian in each corrected attitude block.

When a pose is cloned, let $E$ select its position and attitude errors.
The covariance augmentation is
\begin{equation}
 P_{\mathrm{aug}}=
 \begin{bmatrix}P&PE^\top\\EP&EPE^\top\end{bmatrix}.
 \label{eq:clone-augmentation}
\end{equation}
Retiring a reference removes only the corresponding
six coordinates, retaining the marginal covariance of all remaining states.

\section{Experiments}

\subsection{Experimental Setup and Evaluation}
\begin{figure}[tb!]
    \centering
    \includegraphics[width=\columnwidth]{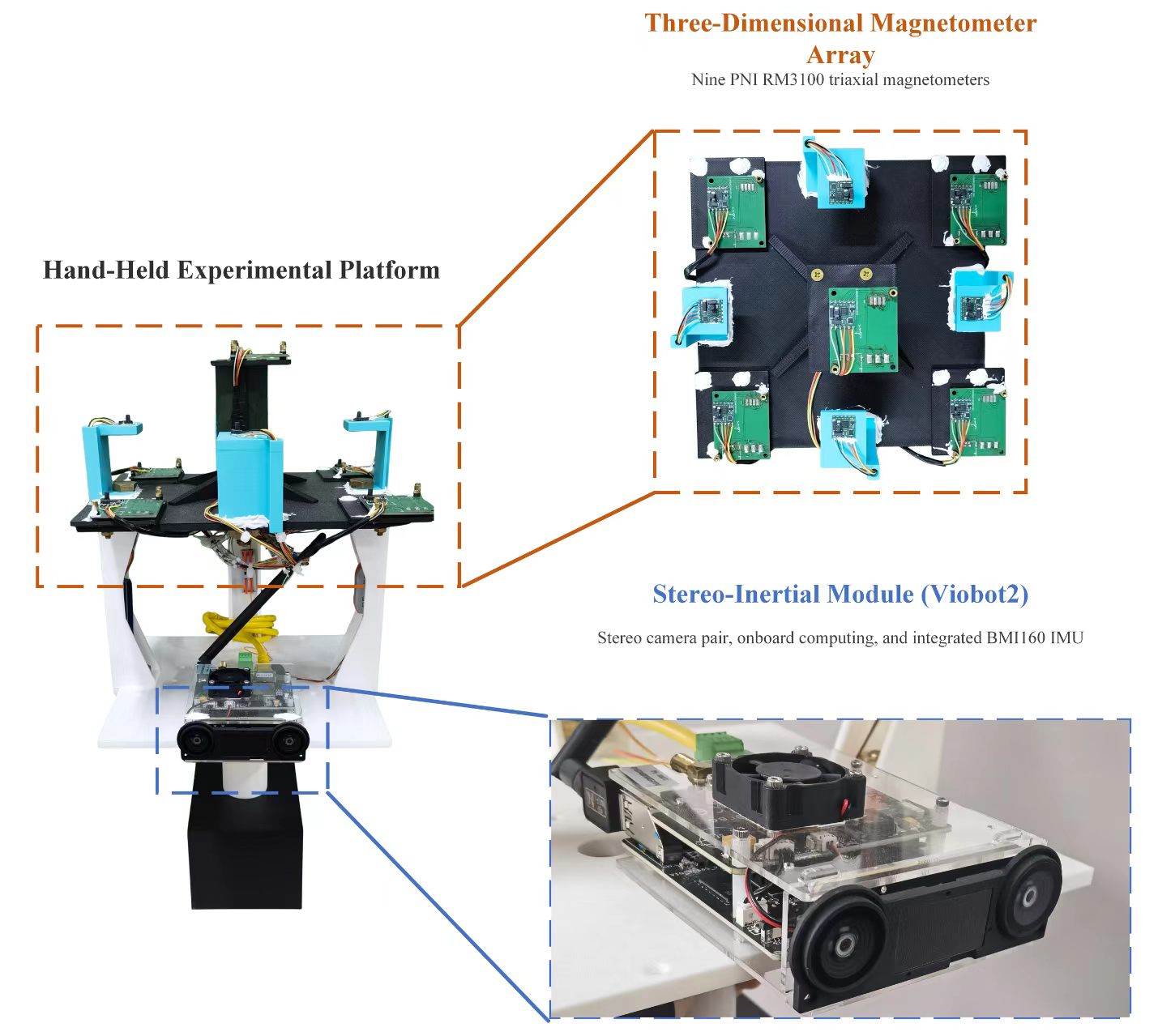}
    \caption{Handheld IMU–magnetometer-camera sensor rig with enlarged views of the three-dimensional nine-magnetometer array and the stereo-inertial
module. The magnetometer array spans $181\times160\times87$~mm.}
    \label{fig:sensor rig}
\end{figure}

We collected data using the handheld sensor rig shown in Fig.~\ref{fig:sensor rig}, which integrates a nine-magnetometer array and a stereo-inertial module. 
The rig uses a Bosch BMI160 IMU and nine PNI RM3100 magnetometers arranged over a $181\times160\times87$~mm array. The stereo camera records $640\times480$-pixel images at 25~Hz with a 60-mm baseline. 
Both the magnetometer array and stereo-inertial module share a common system clock; during acquisition, IMU readings sampled at 200~Hz were interpolated to match the 100-Hz magnetometer timestamps.
In total, three datasets were recorded on two floors of a building and in an underground parking lot, as summarized in Table~\ref{tab:dataset_characteristics}.  Each recording starts and ends at the same location. 
The trajectories and evaluation objectives are described below.
\subsubsection{Multi-loop}
The three-lap square trajectory was collected on the fifth floor of a building, enabling assessment of position tracking along straight segments and heading estimation through 90° turns. Successive laps provide repeated visits
to the same region, allowing us to assess the effect of loop closure in MISLAM.

\subsubsection{Single-loop}
The single-loop trajectory, recorded over nearly two minutes in a large hall on the sixth floor of the same building, allows us to evaluate the magnitude of the accumulated errors in exploring unknown environments.

\subsubsection{Visually degraded}
This dataset was collected in an underground parking lot with insufficient lighting. The experiment evaluates the ability of MISLAM to maintain tracking under poor illumination.

\begin{table}[tb!]
\centering
\caption{Characteristics of the datasets}
\label{tab:dataset_characteristics}
\scriptsize
\setlength{\tabcolsep}{2pt}

\begin{threeparttable}
\begin{tabular}{@{}lccccc@{}}
\hline
Dataset & \shortstack{Duration\\(s)} & \shortstack{Length\\(m)} &
\shortstack{Field magnitude \\std. (snapshot)\tnote{a}  ($\mu$T)} &
\shortstack{Field magnitude\\std.\tnote{b}  ($\mu$T)} &
\shortstack{Visual\\condition} \\
\hline
Multi-loop & 94 & 90 & 1.23 & 5.38 & Good \\
Single-loop & 117 & 90 & 1.52 & 11.46 & Fair \\
Visually degraded & 153 & 120 & 2.55 & 15.57 & Poor \\
\hline
\end{tabular}

\begin{tablenotes}[flushleft]
\scriptsize
\item[a] At each snapshot, we compute the standard deviation of the magnetic-field magnitudes measured by the nine magnetometers; the column reports the median of these values.
\item[b] We first average the nine calibrated magnetic-field vectors at each snapshot, then compute the standard deviation of the magnitude of this mean vector over the dataset.
\end{tablenotes}
\end{threeparttable}
\end{table}

\begin{table}[tb!]
\centering
\begin{threeparttable}
\caption{MISLAM settings for the three datasets}
\label{tab:mislam-settings}
\scriptsize
\setlength{\tabcolsep}{3pt}
\begin{tabularx}{\columnwidth}{@{}Xccc@{}}
\toprule
Setting & Multi-loop & Single-loop & \shortstack{Visually\\degraded} \\
\midrule
Polynomial model degree $n$
    & \multicolumn{3}{c}{2} \\
$\Sigma_{\text{fit}}$ ($\mu$T$^2$) &  \multicolumn{3}{c}{$0.01^2 \cdot I_{3N_L}$ }  \\
\textit{minRefAge}    & \multicolumn{3}{c}{40~ms} \\
\textit{maxRefAge}    & \multicolumn{3}{c}{250~ms} \\
Chi-square test rejection prob. $\alpha_{\mathrm{fit}}$ & \multicolumn{3}{c}{0.01} \\ 
Length scale $\ell$  & \multicolumn{3}{c}{1~m} \\
$N_{\text{\tiny{acc}}}$ & \multicolumn{3}{c}{20} \\
Innovation rejection prob. $\alpha$ & \multicolumn{3}{c}{0.01} \\
\midrule
\multicolumn{4}{@{}l}{\emph{Dataset-specific}} \\
$\Sigma_{L,k}^{\text{\tiny{loc}}}$ ($\mu$T$^2$) & $0.03^2 \cdot I_{3N_L}$  &  $0.03^2 \cdot I_{3N_L}$  & $0.07^2 \cdot I_{3N_L}$ \\
Number of basis functions $M$ & 50 & 100 & 300 \\
$\sigma_b$ & 1 & 1 & 0.8\\
$\sigma_p$ & 50 & 50 & 25\\
$\Sigma_k^{\text{\tiny{GP}}}$ ($\mu$T$^2$) & $0.1^2 \cdot I_{3N_G}$   & $0.8^2 \cdot I_{3N_G}$ & $0.4^2 \cdot I_{3N_G}$ \\
\bottomrule
\end{tabularx}
\end{threeparttable}
\end{table}

We use these datasets to compare MISLAM with MAINS, a state-of-the-art magnetic-inertial odometry~\cite{huang2023mains}, and OpenVINS, a visual-inertial odometry system~\cite{Geneva2020ICRA}. 
Table~\ref{tab:mislam-settings} reports the MISLAM parameters used for each dataset. These parameters were manually tuned based on the estimated trajectories. The local group contains all magnetometers except the central one, which is reserved for the global group.
We evaluate all methods offline and use endpoint closure error as the evaluation metric.

\subsection{Experimental Results}

\begin{figure*}[t!]
    \centering
    \subfloat[Multi-loop]{\includegraphics[width=0.49\textwidth]{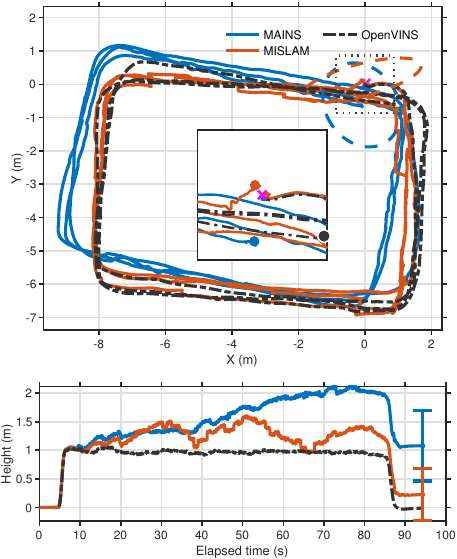}}\hfill
    \subfloat[Single-loop]{\includegraphics[width=0.49\textwidth]{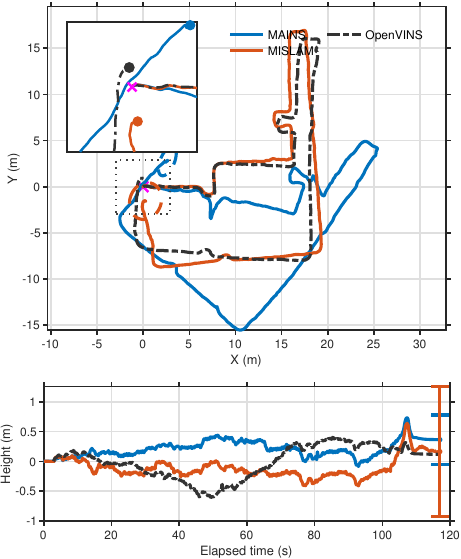}}
    \caption{Trajectory overlays (top) and height estimates (bottom) for the multi-loop and single-loop experiments. Ellipses and error bars indicate $95\%$ uncertainty regions for the final horizontal position estimates and $95\%$ uncertainty intervals for the final height estimates, respectively. The magenta cross marks the starting point.}
    \label{fig:trajectory-comparison}
\end{figure*}

\begin{figure*}[t!]
    \centering
    \subfloat[]{\begin{minipage}[t]{0.49\textwidth}
        \vspace{0pt}
        \centering
        \includegraphics[width=\linewidth]{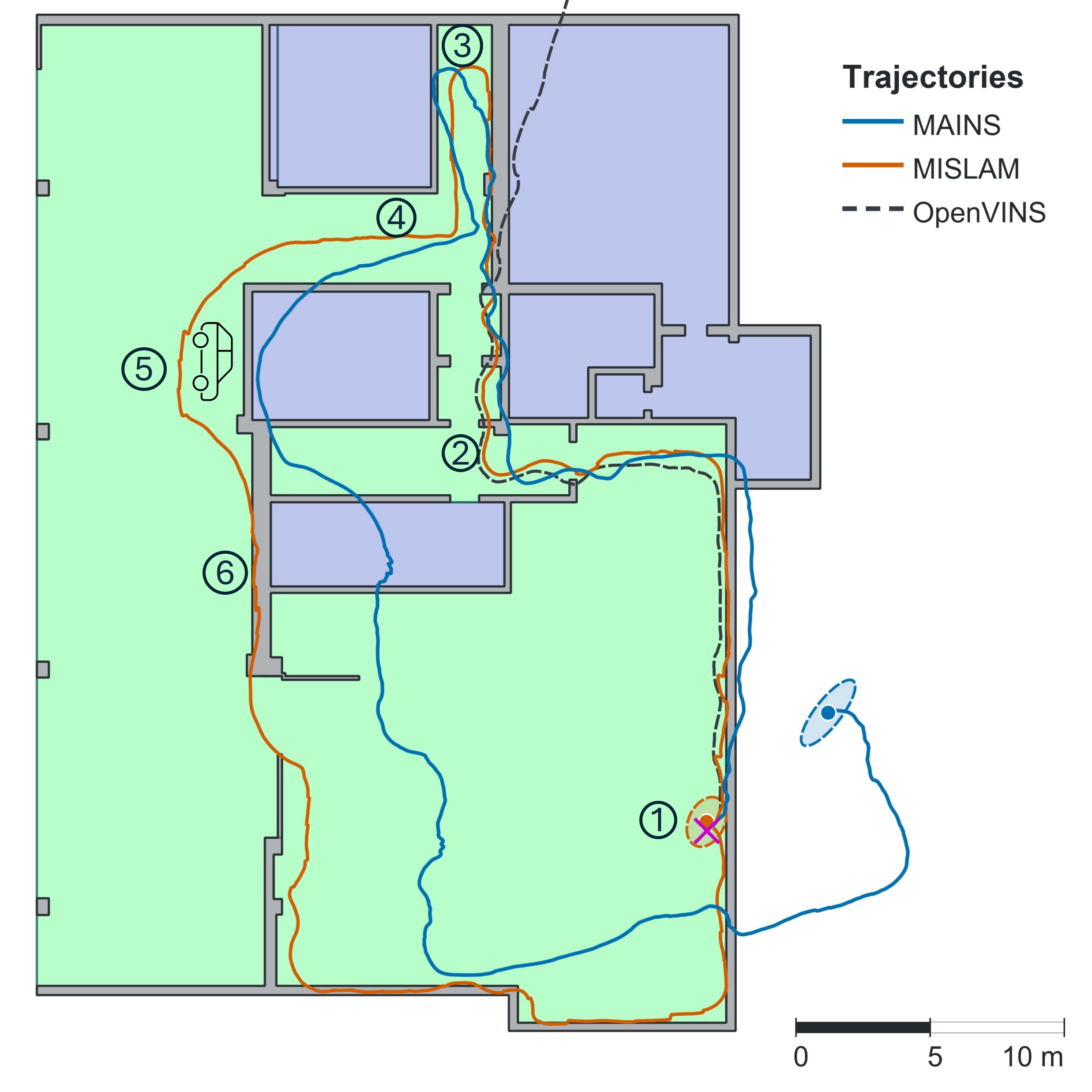}
        \par\medskip
        \includegraphics[width=\linewidth]{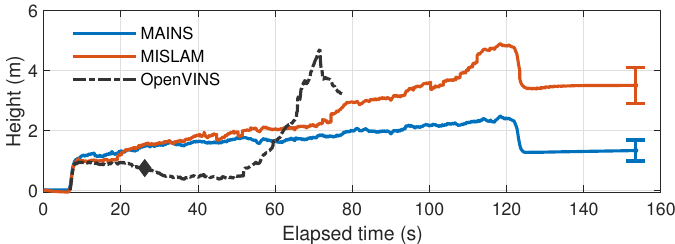}
    \end{minipage}}\hfill
    \begin{minipage}[t]{0.49\textwidth}
        \vspace{0pt}
        \centering
        \subfloat[Location 1]{%
            \includegraphics[width=0.48\linewidth]{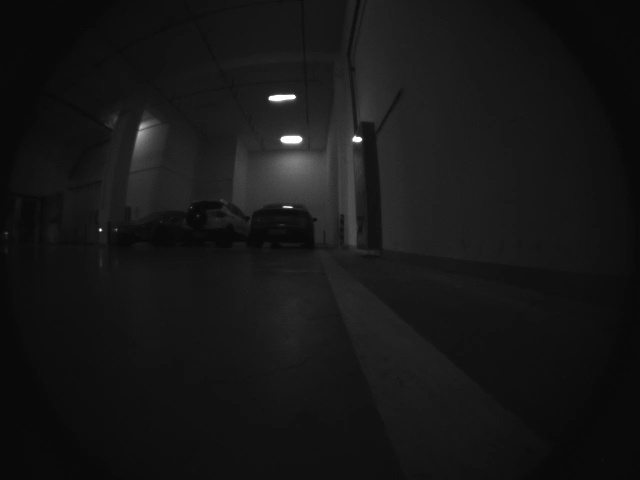}
        }\hfill
        \subfloat[Location 2]{%
            \includegraphics[width=0.48\linewidth]{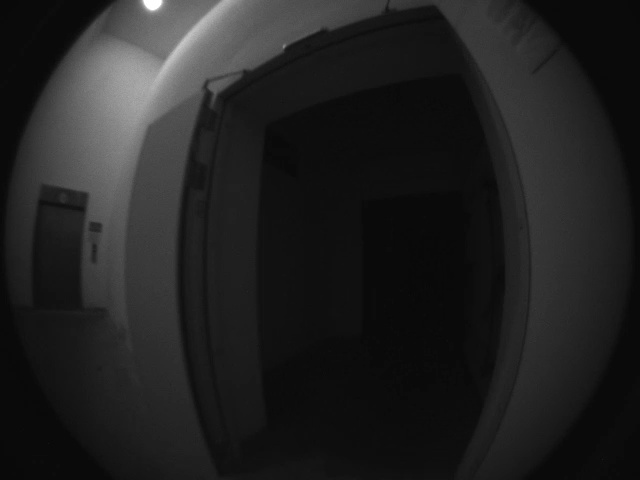}
        }

        \par\medskip
        \subfloat[Location 3]{%
            \includegraphics[width=0.48\linewidth]{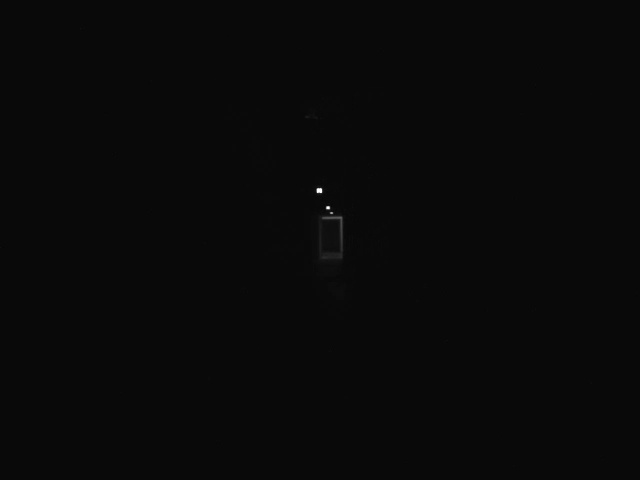}
        }\hfill
        \subfloat[Location 4]{%
            \includegraphics[width=0.48\linewidth]{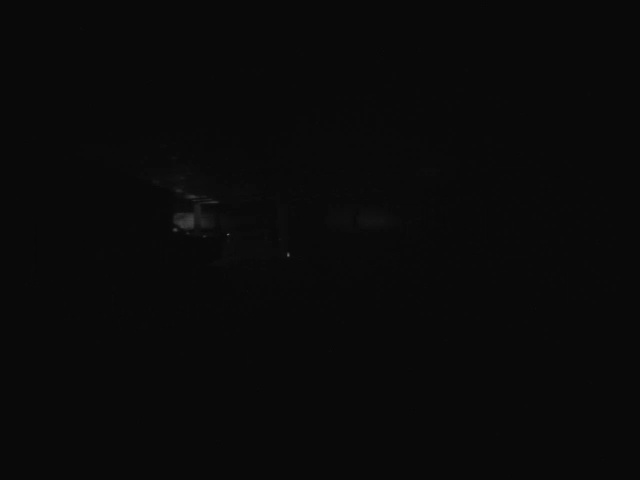}
        }

        \par\medskip
        \subfloat[Location 5]{%
            \includegraphics[width=0.48\linewidth]{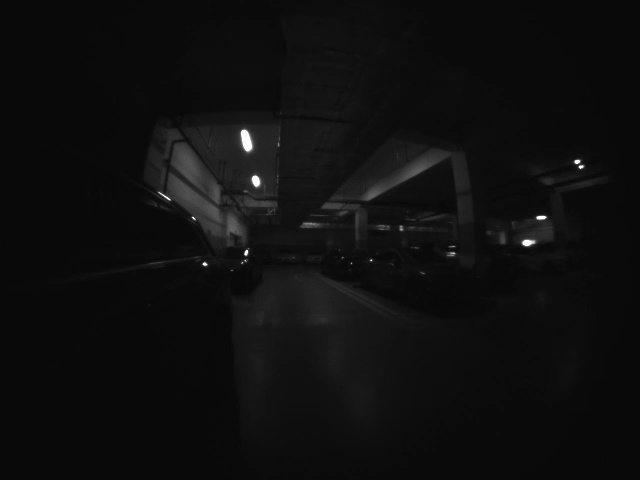}
        }\hfill
        \subfloat[Location 6]{%
            \includegraphics[width=0.48\linewidth]{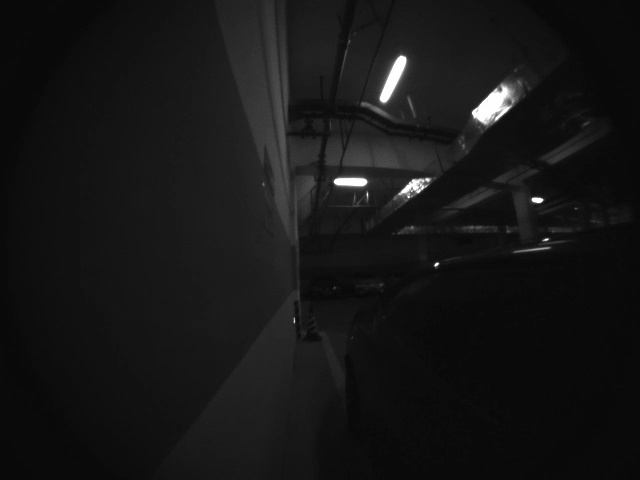}
        }
    \end{minipage}

    \caption{Visually degraded experiment. (a) Trajectory overlay (top) and height estimates (bottom). Scene images (b)--(g) correspond to numbered locations 1--6 in (a), respectively. OpenVINS lost tracking in the poorly lit corridor and terminated around 80~s.}
    \label{fig:visually-degraded-experiment}
\end{figure*}

Table~\ref{tab:experiments} summarizes the endpoint closure errors for the three datasets, while Figs.~\ref{fig:trajectory-comparison} and~\ref{fig:visually-degraded-experiment} show the estimated trajectories and height profiles. MISLAM achieves a lower total closure error than MAINS in all three experiments, with reductions of 78.0\%, 58.9\%, and 45.7\% for the multi-loop, single-loop, and visually degraded datasets, respectively. Its performance relative to OpenVINS depends on the trajectory and visual conditions, as discussed below.

In the multi-loop experiment, MISLAM achieves the smallest total and horizontal closure errors, at 0.274~m and 0.161~m, respectively. The corresponding errors are 1.248~m and 0.628~m for MAINS, and 0.989~m and 0.989~m for OpenVINS. As shown in Fig.~\ref{fig:trajectory-comparison}(a), the MISLAM trajectories from successive laps exhibit less lateral separation than those of MAINS. MISLAM also reduces the vertical closure error from 1.079~m to 0.223~m, although OpenVINS provides the smallest vertical error, at 0.025~m. These observations are consistent with the benefit of retaining a global magnetic map to constrain accumulated drift when revisiting previously explored regions.

In the single-loop experiment, MISLAM reduces the total closure error from 3.831~m for MAINS to 1.576~m. Most of this improvement occurs in the horizontal plane, where the error decreases from 3.813~m to 1.568~m. The trajectory in Fig.~\ref{fig:trajectory-comparison}(b) also follows the overall shape of the OpenVINS estimate more closely than the MAINS trajectory. OpenVINS nevertheless achieves the lowest total closure error, at 0.891~m. The larger MISLAM closure error relative to the multi-loop experiment is consistent with the more limited opportunities to revisit mapped regions during exploration, although the two datasets also differ in their trajectories and magnetic-field characteristics.

The visually degraded experiment illustrates the ability to continue tracking when visual measurements become unreliable. OpenVINS loses tracking in the poorly lit corridor and terminates at approximately 80~s, whereas both magnetic-inertial methods complete the recording. MISLAM achieves a horizontal closure error of 0.332~m, compared with 6.318~m for MAINS, a reduction of 94.7\%. However, the height plot in Fig.~\ref{fig:visually-degraded-experiment}(a) shows substantial drift in MISLAM: its vertical closure error is 3.486~m, exceeding the 1.315~m error of MAINS and dominating its total error of 3.502~m. Thus, the improvement in this experiment is primarily horizontal, and vertical drift remains a limitation. 

\begin{figure*}
    \centering
    \includegraphics[width=\textwidth]{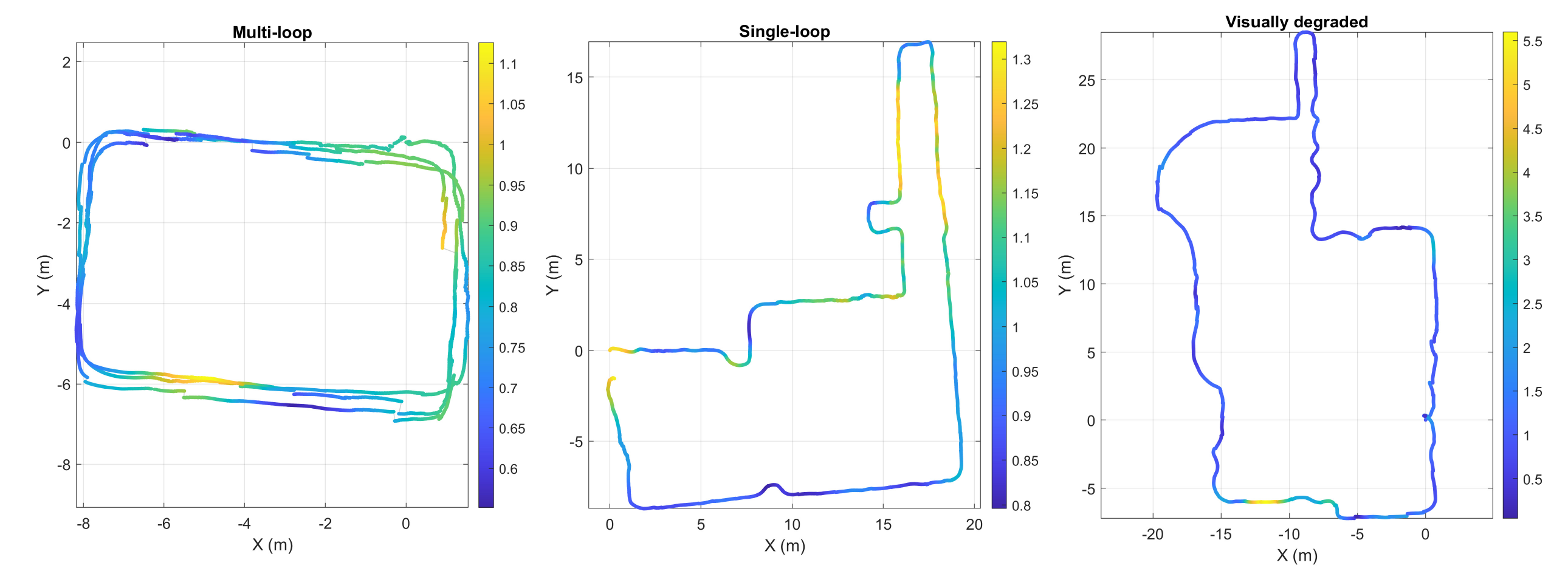}
    \caption{Estimated GP magnetic-field magnitude evaluated along the estimated trajectories and displayed in the horizontal plane. Colors indicate the magnitude of the final posterior mean relative to that of the calibration field. Each panel uses its own color scale.}
    \label{fig:constructed-magnetic-field-map}
\end{figure*}

Finally, Fig.~\ref{fig:constructed-magnetic-field-map} visualizes the estimated magnetic-field magnitude relative to that of the calibration field along the trajectories. These plots provide a qualitative view of the field representation in MISLAM. 
The multi-loop dataset repeatedly covers approximately the same region, allowing MISLAM to reuse previously acquired field information to correct accumulated drift. The resulting pose corrections are visible as discontinuities in the estimated trajectory at global magnetic updates.


\begin{table}[t]
\caption{Endpoint closure errors across three experiments}
\label{tab:experiments}
\centering
\footnotesize
\setlength{\tabcolsep}{3pt}
\begin{tabular}{llrrr}
\hline
Experiment & Algorithm & Total & Horizontal & Vertical \\
           &           & (m)   & (m)        & (m)      \\
\hline
Multi-loop    & MAINS       & 1.248 & 0.628 & 1.079 \\
           & MISLAM    & \textbf{0.274} & \textbf{0.161} & 0.223 \\
           & OpenVINS    & 0.989 & 0.989 & \textbf{0.025} \\
Single-loop & MAINS    & 3.831 & 3.813 & 0.364 \\
           & MISLAM    & 1.576 & 1.568 & 0.163 \\
           & OpenVINS    & \textbf{0.891} & \textbf{0.883} & \textbf{0.119} \\
Visually degraded & MAINS      & 6.453 & 6.318 & \textbf{1.315} \\
           & MISLAM    & \textbf{3.502} & \textbf{0.332} & 3.486 \\
           & OpenVINS$^{\dagger}$ & -- & -- & -- \\
\hline
\end{tabular}
\parbox{\columnwidth}{\footnotesize
$^{\dagger}$OpenVINS terminates early in the visually degraded environment.}
\end{table}

\section{CONCLUSIONS}
This letter presents MISLAM, a tightly coupled magnetic–inertial SLAM system that uses an IMU and a nine-magnetometer array, 
without relying on external odometry or platform-specific motion assumptions. Offline evaluation on three datasets collected with a handheld sensor rig shows that MISLAM reduces total endpoint closure error by 
$45.7\%$–$78.0\%$ relative to MAINS, achieving a total closure error of 0.274~m in the multi-loop experiment. In the visually degraded 
experiment, MISLAM continued tracking throughout the recording after OpenVINS lost tracking and achieved a horizontal closure error of 0.332~m. These results
support the potential of magnetic-inertial SLAM for indoor localization under poor illumination. However, the vertical closure error of 3.486~m in this 
experiment shows that accurate horizontal closure does not necessarily imply accurate three-dimensional localization. Future work will investigate the causes of vertical drift, improve height estimation, explore alternative magnetic-field models, and integrate the system into existing multisensor-fusion frameworks for field robots. 




\bibliographystyle{plain}
\bibliography{refs}

\end{document}